\documentclass[10pt,conference,a4paper]{IEEEtran}
\ifCLASSOPTIONcompsoc
  \usepackage[nocompress]{cite}
\else
  \usepackage{cite}
\fi
\ifCLASSINFOpdf
\else
\fi
\usepackage{xcolor,colortbl,multicol,multirow,mathrsfs,epsfig,graphicx,subfigure,amsmath,amsthm,amsfonts,amssymb,bm,algorithmic,algorithm}
\newtheorem{definition}{Definition}
\begin{document}
%
\title{A constrained clustering based approach for matching a collection of feature sets}
\author{\IEEEauthorblockN{Junchi Yan$^*$}
\IEEEauthorblockA{East China Normal University\\
IBM Research -- China\\
yanjc@cn.ibm.com}
\and
\IEEEauthorblockN{Zhe Ren}
\IEEEauthorblockA{Shanghai Jiao Tong University\\
Shanghai, China\\
sunshinezhe@sjtu.edu.cn}
\and
\IEEEauthorblockN{Hongyuan Zha}
\IEEEauthorblockA{East China Normal University\\
Georgia Institute of Technology\\
zha@cc.gatech.edu}
\and
\IEEEauthorblockN{Stephen Chu}
\IEEEauthorblockA{IBM Watson Research Center\\
Yorktown, New York, USA\\
schu@us.ibm.com}}


%


\maketitle

\begin{abstract}
In this paper, we consider the problem of finding the feature correspondences among a collection of feature sets, by using their point-wise unary features. This is a fundamental problem in computer vision and pattern recognition, which also closely relates to other areas such as operational research. Different from two-set matching which can be transformed to a quadratic assignment programming task that is known NP-hard, inclusion of merely unary attributes leads to a linear assignment problem for matching two feature sets. This problem has been well studied and there are effective polynomial global optimum solvers such as the Hungarian method. However, it becomes ill-posed when the unary attributes are (heavily) corrupted. The global optimal correspondence concerning the best score defined by the attribute affinity/cost between the two sets can be distinct to the ground truth correspondence since the score function is biased by noises. To combat this issue, we devise a method for matching a collection of feature sets by synergetically exploring the information across the sets. In general, our method can be perceived from a (constrained) clustering perspective: in each iteration, it assigns the features of one set to the clusters formed by the rest of feature sets, and updates the cluster centers in turn. Results on both synthetic data and real images suggest the efficacy of our method against state-of-the-arts.
\end{abstract}

%
\IEEEpeerreviewmaketitle
\section{Introduction}
Finding feature correspondence between two or more feature sets \cite{ScottPRSL93} is a fundamental problem in computer vision and pattern recognition, which also relates to operational research as it involves solving assignment problems. As a building block, it facilitates various problems e.g. 3-D reconstruction \cite{YanACCV09}, CAD \cite{NiuCAD15,niu2015applying}, visual tracking \cite{LeePR93}, common object discovery \cite{ChoCVPR15}, among a considerable amount of applications.

Most existing methods for feature correspondence have been devoted into the pairwise case i.e. matching two feature sets one time \cite{MacielPAMI03}. Nevertheless it is far more common in real-world problems where a batch of feature sets are involved.

This paper focuses on the problem that given multiple (more than two) feature sets, how to find their \emph{one-to-one} correspondences based on the unary node-wise attributes without considering the higher-order compatibility between nodes. For the classical two-set setting, the problem usually can be transformed to a linear assignment problem whose global optimum can be found in polynomial time using the Hungarian method \cite{munkres1957algorithms}. While this paper explores the setting where a collection feature sets are involved for feature matching.
\section{Related Work}
We mention a few of relevant concepts about the correspondence problem, mainly in the context of computer vision.
\subsection{Point registration and feature matching}
For point registration, often a parametric transformation is assumed \cite{ChuiCVIU03,YanAAAI15P} which serves as a regularization or prior term in the objective to account for the motion in an image sequence. The registration problem usually involves two `chicken-and-egg' steps: i) finding correspondence from two point sets; ii) estimating the parameters of the transformation based on the correspondence. For instance, based on an initial correspondence, the iterative closest point (ICP) methods \cite{ChenIVC92,ZhangIJCV94} iterate between finding the correspondence via nearest neighbor and updating the transformation with the least square error. Many other methods, such as the Robust Point Matching (RPM) \cite{ChuiCVIU03} and the Coherent Point Drift (CPD) method \cite{MyronenkoPAMI10} are designed to perform pairwise registration. In contrast, the multi-view methods e.g. \cite{YanAAAI15P} aim to finding the correspondence and geometric transform over a batch of point sets.

Similar to the problem considered in this paper, there are several methods that only explore the unary feature attributes associated with the nodes without imposing geometrical priors. Two examples are \cite{ZengECCV12,YuPR16}. The former formulates the feature matching among multiple sets as a matrix low rank sparsity decomposition task; the latter proposes an optimization algorithm to find consistent correspondences over feature sets.
\subsection{Graph matching}
In contrast to point registration, there is no parametric transformation imposed in graph matching \cite{ConteIJPRAI04,FoggiaIJPRAI14} thus we call it a non-parametric model. Moreover, compared with feature matching, graph matching additionally incorporates the edge attribute which can refer to second-order \cite{TianECCV12}, or even higher-order \cite{YanCVPR15,NgocCVPR15} geometrical information for matching. This lifts the order of the matching problem, and a quadratic assignment programming formulation is derived \cite{LoiolaEJOR07,LeordeanuNIPS09} which in general is known NP-hard and only a few special cases can be solved in polynomial time \cite{AflaloPNAS15}, such as planar graph \cite{EppsteinSODA95}, bounded valence graph \cite{luks1982isomorphism}, and tree structure \cite{aho1989code}. Note there are several state-of-the-art multiple graph matching methods \cite{YanICCV13,YanTIP15,YanECCV14,YanPAMI16,ShiCVPR13,ShiCVPR16} by which the multiple feature sets matching problem can be viewed as a special case in their model by ignoring the edge attributes, however, as will be shown later in this paper, our method specifically leverages the special structure of the problem for the unary feature that enables the proposed clustering \cite{YanYanPAMI16} based approach. These methods will also be compared in our tests.

Note the incorporation of higher-order information or transform prior as done in graph matching or point registration may be beneficial in certain practical tasks, yet this is a problem in parallel as they use different assumptions and information.
\subsection{Matching cost modeling}
It is theoretically proved in \cite{BunkePAMI99} that the graph edit costs is critical to the graph edit distance based methods. In complex tasks a manual procedure for cost setting is difficult, or even impossible to apply. To address this issue, \cite{neuhaus2007automatic} aims to learn the edit cost by a probabilistic framework, to reduce the intra-class edit distance and increase the inter-class one. Alternatively, \cite{neuhaus2005self} proposes to use self-organizing maps to learn the edit cost. For graph matching, recent work leverage various machine leaning algorithms for computing the optimal affinity matrix~\cite{LeordeanuIJCV12,ChoICCV13}, and these methods in general fall into either supervised~\cite{ChoICCV13}, unsupervised~\cite{LeordeanuIJCV12}, or semi-supervised~\cite{LeordeanuICCV11} learning paradigms, based on to what extent the supervision information is used. The cost modeling is orthogonal to this paper and we assume the cost is given.
\subsection{Correspondence consistency over multiple node sets}
\begin{figure}[t]
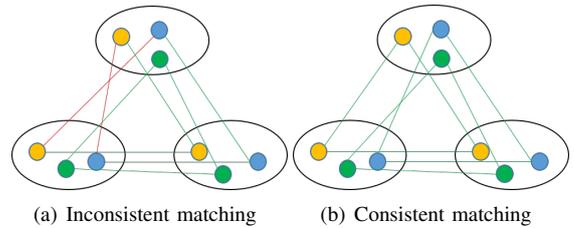

\centering
\subfigure[Inconsistent matching]{\label{fig:inconsistent}
\includegraphics[width=0.2\textwidth]{{{pic/inconsistent}}}}\hspace{-5pt}
\subfigure[Consistent matching]{\label{fig:consistent}
\includegraphics[width=0.2\textwidth]{{{pic/consistent}}}}\vspace{-5pt}
\caption{Illustration of globally consistent correspondence for three feature sets $S_1$, $S_2$ and $S_3$: feature points in the same color relate to the same entity.}
\label{fig:consistency}
\end{figure}

Matching consistency around multiple node sets is a widely recognized concept such as \cite{YanICCV13,YanPAMI16}. The key idea is illustrated in Fig.\ref{fig:consistency}: given the pairwise correspondence set $\mathbb{X}=\{\textbf{X}_{ij}\}_{i=1,j=i+1}^{N-1,N}$ independently computed by a two-set matching solver  from $N$ feature sets -- we call $\mathbb{X}$ as \emph{matching configuration} in line with \cite{YanPAMI16} where $\textbf{X}_{ij}$ is the node correspondence permutation matrix for feature set $S_i$ and $S_j$\footnote{Specifically, the element $x^{st}_{ij}$ in $\textbf{X}_{ij}$ denotes the $s$th node in feature set $S_i$ corresponds to the $t$th node in feature set $S_j$ if $x^{st}_{ij}=1$, otherwise $x^{st}_{ij}=0$.}, there often exists the inconsistency over different node correspondence transitive paths. Now we further introduce two definitions concerning consistency as presented in \cite{YanPAMI16,YanTIP15} (we slightly rewrite them to better fit to our problem).
\begin{definition}\label{def:single_consist}
Given $N$ feature sets $\{\mathcal{S}_k\}_{k=1}^N$ and the pairwise matching configuration $\mathbb{X}=\{\textbf{X}_{ij}\}_{i=1,j=i+1}^{N-1,N}$, the unary consistency of feature set $\mathcal{S}_k$ is defined as $C_u(k,\mathbb{X})=1-\frac{\sum_{i=1}^{N-1}\sum_{j=i+1}^{N}\|\textbf{X}_{ij}-\textbf{X}_{ik}\textbf{X}_{kj}\|_F/2}{nN(N-1)/2}\in(0,1]$, where $\|\cdot\|_F$ is the Frobenious norm.
\end{definition}
\begin{definition}\label{def:pairwise_consist}
Given feature sets $\{\mathcal{S}_k\}_{k=1}^N$ and matching configuration $\mathbb{X}$, for any pair $\mathcal{G}_i$ and $\mathcal{G}_j$, the pairwise consistency is defined as $C_{p}(\textbf{X}_{ij},\mathbb{X})=1-\frac{\sum_{k=1}^{N}\|\textbf{X}_{ij}-\textbf{X}_{ik}\textbf{X}_{kj}\|_F/2}{nN}\in(0,1]$.
\end{definition}
where $n$ is the number of feature points in each feature set. In the next section, we will present our method where the above two definitions will be used.
\section{Proposed Method}
For two-set feature matching with set $S_1$ and $S_2$, by merely taking the unary attributes into consideration, the matching problem can be formulated as a linear assignment problem, for $n_1\leq n_2$ without loss of generality:
\begin{gather}\label{eq:linear assignment}
\min_{x_{ij}}\sum_{i=1}^{n_1}\sum_{j=1}^{n_2} c_{ij}x_{ij}\\\notag
\text{s.t.}\quad \{\sum_{j=1}^{n_2}x_{ij}=1\}_{i=1}^{n_1}, \{\sum_{i=1}^{n_1}x_{ij}\leq1\}_{j=1}^{n_2},x_{ij} \in\{0,1\}\notag
\end{gather}
Or in a more compact form:
\begin{gather}\label{eq:linear assignment_vec}
\min_{\textbf{X}}\text{vec}(\textbf{C})^T\text{vec}(\textbf{X})\\\notag
s.t. \quad \textbf{X}\textbf{1}_{n_2}\leq \textbf{1}_{n_1}\quad \textbf{1}^T_{n_1}\textbf{X}=\textbf{1}^T_{n_2} \quad \textbf{X} \in \{0,1\}^{n_1\times n_2}
\end{gather}
The operation $\text{vec}(\cdot)$ stacks the columns of the input matrix into a column vector. The superscript $T$ denotes the transformation of a given matrix or vector, and $c_{ij}$ is the element in matrix $\textbf{C}$. Here $\textbf{X}$ is a partial permutation matrix.

The cost matrix $\textbf{C}$ can be modeled in different ways based on the applications. In line with \cite{ScottPRSL93}, we compute it by an exponential cost of the feature vector distance. Formally, let the feature vectors of feature set $S_i$ be $\textbf{F}_i=[\textbf{f}_1,\textbf{f}_2,\ldots,\textbf{f}_n]$ where $\{\textbf{f}_t\}_{t=1}^n\in\mathcal{R}^{D}$ is the feature vector of feature $t$ in $S_i$, thus $\textbf{F}_i\in\mathcal{R}^{D\times n}$. Then the cost between assigning feature $s$ in one feature set to feature $t$ in the other is computed by:
\begin{gather}\label{eq:cost_model}\notag
\left\{c_{st}=\exp\left(-{\|\textbf{f}_{s}-\textbf{f}_{t}\|_2}/\sigma^2D\right)\right\}_{s=1,t=1}^{n_1,n_2}
\end{gather}
we set the scale parameter $\sigma^2$ to $0.15$ in this paper which is found insensitive to the performance, and $\|\cdot\|_2$ is the $\ell_2$ norm.

The above linear assignment problem has been well studied in early years and polynomial solvers that can attain global optimum are devised such as the Hungarian method \cite{munkres1957algorithms} and its Jonker-Volgenant alternative \cite{JonkerComputing87}.

As a common preprocessing step, we replace the inequality constraint with the equality constraint as follows:
\begin{gather}\label{eq:linear assignment_vec}
\min_{\textbf{X}}\text{vec}(\textbf{C})^T\text{vec}(\textbf{X})\\\notag
s.t. \quad \textbf{X}\textbf{1}_{n_2}=\textbf{1}_{n_1}\quad \textbf{1}^T_{n_1}\textbf{X}=\textbf{1}^T_{n_2} \quad \textbf{X} \in \{0,1\}^{n_1\times n_2}
\end{gather}
\begin{figure}[t]
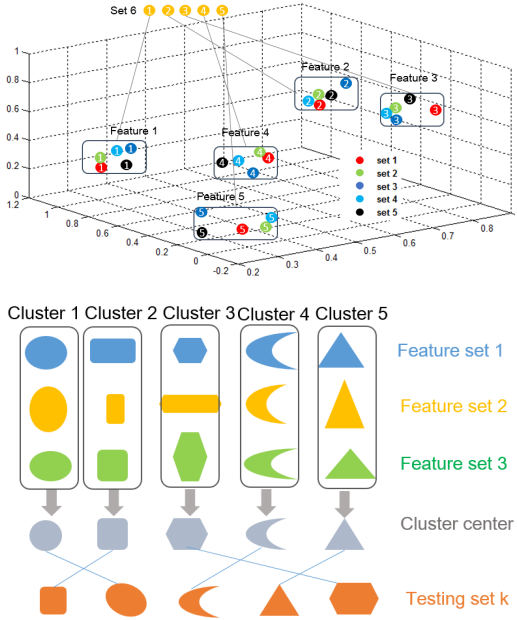

\centering
\subfigure{\label{fig:cluster}
\includegraphics[width=0.37\textwidth]{{{pic/cluster}}}}
\subfigure{\label{fig:cluster_ill}
\includegraphics[width=0.37\textwidth]{{{pic/cluster_ill}}}}\vspace{-5pt}
\caption{\textbf{Top}: illustration of the clustering distribution in a 3-dimension space. In this illustration, there are 5 feature sets with 5 features in each of the set. Different colors correspond to different feature sets. The feature points of the 6th feature set are assigned to the clusters based on their distance to the cluster centers which can be computed using the Hungarian methods. \textbf{Bottom}: a more intuitive illustration of our method matchCluster whereby different shapes denote clusters and same color indicate one feature set.}
\label{fig:assignment_ill}
\end{figure}
This can be fulfilled by adding dummy nodes to feature set $S_1$ of smaller size (i.e. adding slack variables to the assignment matrix and augment the affinity matrix by zeros) for $n_1\neq n_2$, then $n_1=n_2=n$. This is a standard technique from linear programming and is adopted as a robust means by the literature e.g. \cite{YanTIP15,WongPAMI85,YanPAMI16} to handle unmatchable nodes.

However, directly applying the Hungarian method to obtain the mathematical global optimum may not lead to the perfect matching in real-world problems. There are two reasons for the existence of such an ambiguity: i) modeling the cost/affinity function is non-trivial and it is difficult to fit a score function fully unbiased with the matching accuracy; ii) the noises further make the correlation between matching accuracy and score function more biased, especially only two feature sets are given. In contrast, a collection of feature sets is expected to provide additional global information to help disambiguate the local noises by information fusion.

As a result, we consider the multiple feature set matching problem as a clustering task in the feature space: each feature correspondence across the sets can be viewed as one cluster comprised of the corresponding feature points from each set. From this perspective, the matching problem can be viewed as a procedure to group each feature node in each set into different clusters. Different from the classical clustering problem, there is one combinatorial constraint imposed on the clustering procedure: any two feature nodes in one set cannot be assigned to the same cluster since we assume one-to-one node correspondence. Hence traditional clustering methods such as k-mean cannot be directly applied.
\begin{algorithm}[]
\small
  \caption{Fast constrained clustering based multiple feature sets matching (\textbf{matchCluster$^{\text{fast}}$})}
  \label{alg:matchCluster_fast}
  \begin{algorithmic}[1]
    \REQUIRE features matrix of each feature set: $\{\textbf{F}_{i}\}_{i=1}^{N}$;
      \STATE Randomly choose a reference feature set $\textbf{F}_{r}$ and compute two-set correspondences $\{\textbf{X}_{ri}\}_{i=1,i\neq r}^N$ by the Hungarian method;
      \STATE Compute the aligned feature matrix by $\{\textbf{F}'_{i}=\textbf{F}_{i}\textbf{X}_{ri}\}_{i=1,\neq r}^N$;
    \FOR {$k=1:N, k\neq r$}
    \STATE Compute the mean feature matrix $\textbf{F}_{c}=\sum_{i=1,\neq k}^N{\textbf{F}'_i}/(N-1)$;
    \STATE Update $\textbf{X}_{rk}$ by computing the correspondence between $\textbf{F}_{c}$ and $\textbf{F}'_{k}$ via the Hungarian method;
    \ENDFOR
  \end{algorithmic}
\end{algorithm}
\begin{algorithm}[]
\small
  \caption{Constrained clustering based multiple feature sets matching (\textbf{matchCluster})}
  \label{alg:matchCluster}
  \begin{algorithmic}[1]
    \REQUIRE features matrix of each feature set: $\{\textbf{F}_{i}\}_{i=1}^{N}$;
      \STATE Compute the two-set feature correspondences $\mathbb{X}=\{\textbf{X}_{ij}\}_{i,j=1}^N$ for each pair of sets independently by the Hungarian method;
      \STATE Compute unary feature set consistency $\{C_u(k,\mathbb{X})\}_{k=1}^N$ by Def.\eqref{def:single_consist}, and choose the reference $S_r=\max_{r}C_u(r,\mathbb{X})$;
      \STATE Compute pairwise consistency $\{C_{p}(\textbf{X}_{ri},\mathbb{X})\}_{i=1,\neq r}^N$ by Def.\eqref{def:pairwise_consist};
      \STATE Compute the aligned feature matrix by $\{\textbf{F}'_{i}=\textbf{F}_{i}\textbf{X}_{ri}\}_{i=1,\neq r}^N$;
    \FOR {set $S_k$ in descending order by $\{C_{p}(\textbf{X}_{rk},\mathbb{X})\}_{k=1,\neq r}^N$}
    \STATE Compute the mean feature matrix $\textbf{F}_{c}=\sum_{i=1,\neq k}^N{\textbf{F}'_i}/(N-1)$;
    \STATE Update $\textbf{X}_{rk}$ by computing the correspondence between $\textbf{F}_{c}$ and $\textbf{F}'_{k}$ via the Hungarian method;
    \ENDFOR
  \end{algorithmic}
\end{algorithm}
To address this constrained clustering problem, we propose an iterative algorithm as described in Alg.\ref{alg:matchCluster_fast}: we first generate a group of initial two-set feature correspondences: $\{\textbf{X}_{ri}\}_{i=1,\neq r}^N$ where $S_r$ is a reference feature set. Using this reference feature set $\textbf{F}_{r}$ and the correspondences: $\{\textbf{X}_{ri}\}_{i=1,\neq r}^N$, we can obtain a series of aligned feature matrix: $\textbf{F}'_{1},\textbf{F}'_{2},\dots,\textbf{F}'_{N}$ by $\textbf{F}'_{i}=\textbf{F}_{i}\textbf{X}_{ri}$. Then we choose a certain feature set $\textbf{F}_{k}$, and compute the mean of the rest feature sets: $\textbf{F}_{c}=\sum_{i=1,\neq k}^N{\textbf{F}'_i}/(N-1)$. Then we compute the node-to-node cost between $\textbf{F}_{c}$ and $\textbf{F}_{k}$, based on which we employ the Hungarian method to compute their correspondence $\textbf{X}_{ck}$, and we set $\textbf{X}_{rk}=\textbf{X}_{ck}$. As the iteration continues, we update $\textbf{X}_{rk}$ for $k=1,2,\ldots,N; k\neq r$ alternatively by fixing the other $\textbf{X}_{rt}, t\neq k$. The above assignment procedure in one iteration is illustrated in Fig.\ref{fig:assignment_ill}.
\begin{figure}[tb]
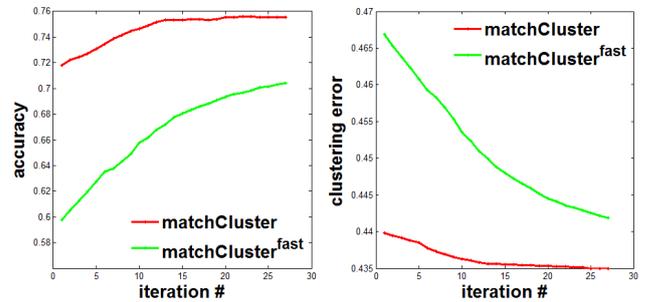

\centering
\subfigure{\label{fig:acc_curve}
\includegraphics[width=0.23\textwidth]{{{pic/acc_curve}}}}\hspace{-5pt}
\subfigure{\label{fig:err_curve}
\includegraphics[width=0.23\textwidth]{{{pic/err_curve}}}}\vspace{-5pt}
\caption{Average accuracy and clustering error curves as a function of the iteration number by matchCluster and matchCluster$^{\text{fast}}$ on the CMU Hotel sequence test by 20 tests with 28 feature sets and 20 features per each trial.}
\label{fig:curve}
\end{figure}
We term Alg.\ref{alg:matchCluster_fast} by a superscript `fast' because it is initialized by a linear number of two-set matchings regarding $N$. However, as discussed in a similar situation in \cite{YanTIP15}, since our method is an alternating updating procedure, the final solution can be sensitive to the quality of the initial point. Moreover, the updating order may also influence the iteration path.

There is one useful observation in \cite{YanPAMI16,YanTIP15} that matching accuracy, though not observable in testing stage, is correlated with the pairwise consistency $\{C_{p}(\textbf{X}_{ri},\mathbb{X})\}_{i=1,\neq r}^N$ as defined in Def.\eqref{def:pairwise_consist}. Moreover, the most consistent feature set by Def.\eqref{def:single_consist} is expected to generate the most accurate initial matching set $\{\textbf{X}_{ri}\}_{i=1,\neq r}^N$ from $\mathbb{X}$. Thus we adopt this strategy and describe the improved algorithm in Alg.\ref{alg:matchCluster}. While the expense is the additional time cost for computing the whole $\mathbb{X}$ with $O(N^2)$ times of pairwise Hungarian method and the computing of $\{C_u(k,\mathbb{X})\}_{k=1}^N$ and $\{C_{p}(\textbf{X}_{ri},\mathbb{X})\}_{i=1,\neq r}^N$. One illustration of the usefulness of the above strategy is plotted in Fig.\ref{fig:curve} in comparison with a random selection (matchCluster$^{\text{fast}}$).
\section{Experiment}
\subsection{Test data settings and compared methods}
\subsubsection{Synthetic feature set matching}
The random graph test provides a controlled setting to evaluate the performance of our methods and other peer methods. For each trial, a reference feature set with $n_{i}$ nodes is created by assigning a random attribute vector of dimension $D$ to each of its nodes. In our test, we set $D=5$ for all synthetic tests. The attributes are uniformly sampled from the interval $[0,1]$. Then the `perturbed' feature sets are created by adding a Gaussian deformation disturbance to the attributes $q_{id}^r$, which is sampled from $N(0,\varepsilon)$ i.e. $q_{id}^p=q_{id}^r+N(0,\varepsilon)$ where the superscript `p' and `r' denotes for `perturb' and `reference' respectively. Optionally, each `perturbed' feature set is added by $n_{o}$ outliers sampled from the same distribution as inliers.
\subsubsection{CMU hotel sequence}
The CMU motion sequences (http://vasc.ri.cmu.edu//idb/html/motion/) are widely used for feature matching \cite{MacielPAMI03} and graph matching \cite{YanCVPR15,YanICCV13,YanTIP15,YanPAMI16} etc. Here we use it to test feature matching solvers. In our experiment, we focus on the \emph{hotel} sequence (111 frames) because matching the other sequence \emph{house} by different methods always produces the perfect matching results. We use 15, 20, 25 feature points from each frame respectively, where the unary feature vector is computed by the SIFT descriptor \cite{LoweICCV99}.
\subsubsection{Willow-ObjectClass}
This dataset released in \cite{ChoICCV13} is constructed using images from Caltech-256 and PASCAL VOC2007. Each object category contains different number of images: 109 Face, 50 Duck, 66 Wine bottle, 40 Motorbike, and 40 Car images. For each image, 10 feature points are manually labeled on the target object. We focus on the testing on the face category because the matching accuracy for other images is much lower (below 0.25) if only unary SIFT feature is considered. This is because the objects are from different angles, different textures and colors, as well as different poses.
\subsubsection{Comparing methods}
We compare several state-of-the-art general multi-view matching methods including matchOpt \cite{YanICCV13,YanTIP15}, matchSync\cite{PachauriNIPS13}, Composition based Affinity Optimization (CAO-) and its consistency-enforcing variant CAO \cite{YanECCV14,YanPAMI16}. In addition, we also involve the baseline two-set matching by the Hungarian method. There are two cases for applying the Hungarian method: i) Hung$^{\text{pair}}$ which performs the Hungarian method independently on each pair of the feature sets. This directly produces the whole matching configuration $\mathbb{X}$; ii) randomly select one feature set $S_r$, and preform the Hungarian method independently on each of the other feature sets with it. Then the pairwise matchings are computed by $\textbf{X}_{ij}=\textbf{X}_{ir}\textbf{X}_{rj}$. Note the former has a quadratic $O(N^2)$ time cost in terms of the number of feature sets $N$, and the latter is in linear with $N$. For our methods, we also test two versions: i) matchCluster which decides the base feature set and alternating order by the consistency in definition Def.\eqref{def:single_consist} and Def.\eqref{def:pairwise_consist} respectively; ii) the fast version matchCluster$^{\text{fast}}$ that randomly sets the base feature set and  alternating order. Note we do not include any point registration method such as \cite{ChuiCVIU03,MyronenkoPAMI10,YanAAAI15P} in our evaluation because we focus on the non-parametric node correspondences and impose no parametric transformation prior nor regularization for fair comparison.
\subsection{Results and discussion}
\begin{figure*}[t]
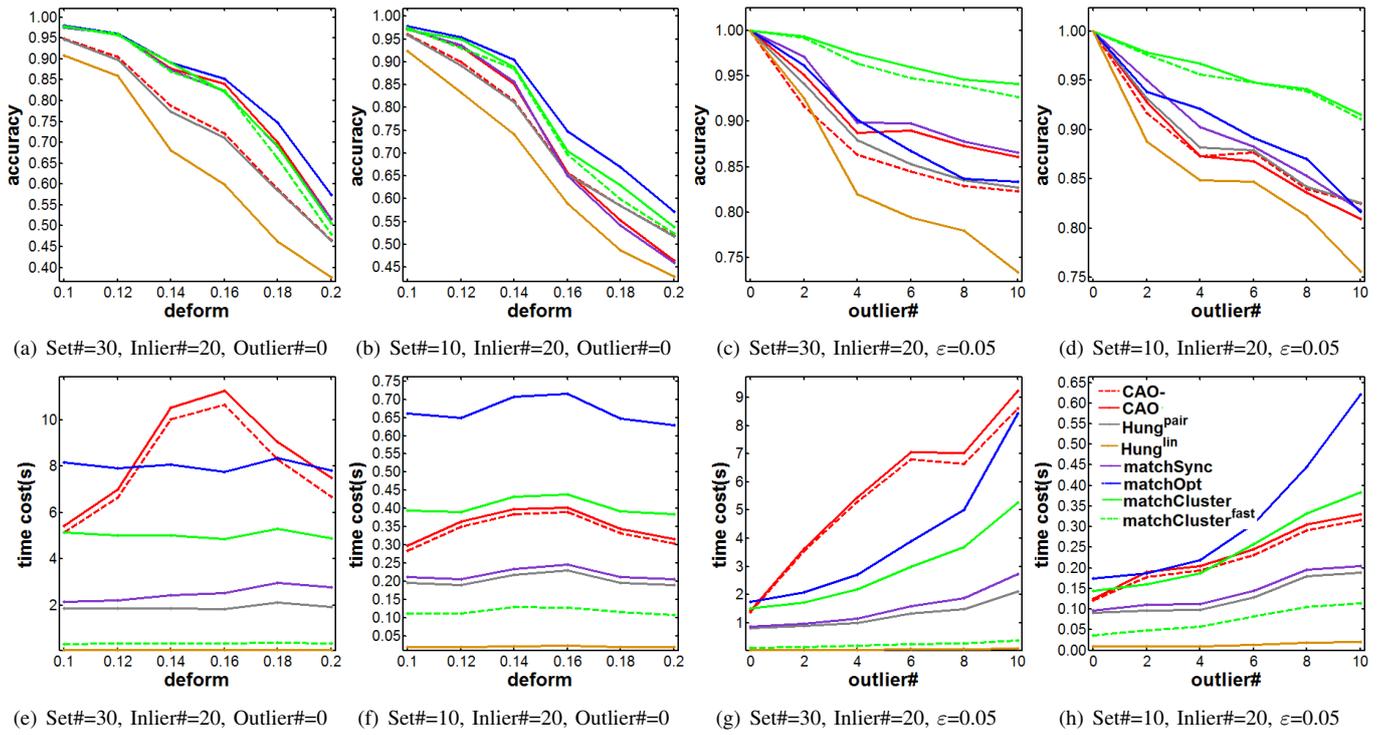

\centering
\subfigure[Set\#=30, Inlier\#=20, Outlier\#=0]{\label{fig:random_deform_acc_set=30}
\includegraphics[width=0.245\textwidth]{{{pic/performance_random_deform_RRWM_RRWM_respect_0.10_0.20_formal_acc}}}}\hspace{-5pt}
\subfigure[Set\#=10, Inlier\#=20, Outlier\#=0]{\label{fig:random_deform_acc_set=10}
\includegraphics[width=0.245\textwidth]{{{pic/performance_random_deform_RRWM_RRWM_respect_0.10_0.20_set=10_acc}}}}\hspace{-5pt}
\subfigure[Set\#=30, Inlier\#=20, $\varepsilon$=0.05]{\label{fig:random_outlier_acc_set=30}
\includegraphics[width=0.245\textwidth]{{{pic/performance_random_outlier_RRWM_RRWM_respect_0_10_formal_acc}}}}\hspace{-5pt}
\subfigure[Set\#=10, Inlier\#=20, $\varepsilon$=0.05]{\label{fig:random_outlier_acc_set=10}
\includegraphics[width=0.245\textwidth]{{{pic/performance_random_outlier_RRWM_RRWM_respect_0_10_set=10_acc}}}}\vspace{-5pt}
\subfigure[Set\#=30, Inlier\#=20, Outlier\#=0]{\label{fig:random_deform_tim_set=30}
\includegraphics[width=0.245\textwidth]{{{pic/performance_random_deform_RRWM_RRWM_respect_0.10_0.20_formal_tim}}}}\hspace{-5pt}
\subfigure[Set\#=10, Inlier\#=20, Outlier\#=0]{\label{fig:random_deform_tim_set=10}
\includegraphics[width=0.245\textwidth]{{{pic/performance_random_deform_RRWM_RRWM_respect_0.10_0.20_set=10_tim}}}}\hspace{-5pt}
\subfigure[Set\#=30, Inlier\#=20, $\varepsilon$=0.05]{\label{fig:random_outlier_tim_set=30}
\includegraphics[width=0.245\textwidth]{{{pic/performance_random_outlier_RRWM_RRWM_respect_0_10_formal_tim}}}}\hspace{-5pt}
\subfigure[Set\#=10, Inlier\#=20, $\varepsilon$=0.05]{\label{fig:random_outlier_tim_set=10}
\includegraphics[width=0.245\textwidth]{{{pic/performance_random_outlier_RRWM_RRWM_respect_0_10_set=10_tim}}}}\vspace{-5pt}
\caption{Matching accuracy and run-time as the disturbance (deformation, number of outliers) vary on the synthetic test.}
\label{fig:random_vary_noise}
\end{figure*}

\begin{figure*}[t]
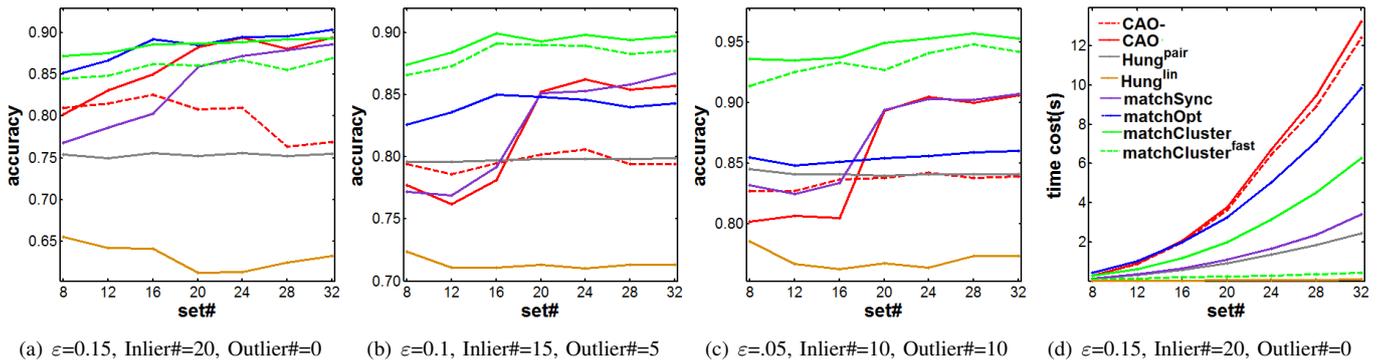

\centering
\subfigure[$\varepsilon$=0.15, Inlier\#=20, Outlier\#=0]{\label{fig:random_grh_cnt_acc1}
\includegraphics[width=0.245\textwidth]{{{pic/performance_random_deform_RRWM_RRWM_respect_def=0.15_out=0_acc}}}}\hspace{-5pt}
\subfigure[$\varepsilon$=0.1, Inlier\#=15, Outlier\#=5]{\label{fig:random_grh_cnt_acc2}
\includegraphics[width=0.245\textwidth]{{{pic/performance_random_outlier_RRWM_RRWM_respect_def=0.1_out=5_acc}}}}\hspace{-5pt}
\subfigure[$\varepsilon$=.05, Inlier\#=10, Outlier\#=10]{\label{fig:random_grh_cnt_acc3}
\includegraphics[width=0.245\textwidth]{{{pic/performance_random_outlier_RRWM_RRWM_respect_def=0.05_out=10_acc}}}}\hspace{-5pt}
\subfigure[$\varepsilon$=0.15, Inlier\#=20, Outlier\#=0]{\label{fig:random_grh_cnt_tim1}
\includegraphics[width=0.245\textwidth]{{{pic/performance_random_deform_RRWM_RRWM_respect_def=0.15_out=0_tim}}}}\vspace{-5pt}
\caption{Matching accuracy and run-time under different disturbance settings, as the number of feature sets vary on the synthetic test.}
\label{fig:vary_graph_cnt}
\end{figure*}

\begin{figure*}[t]
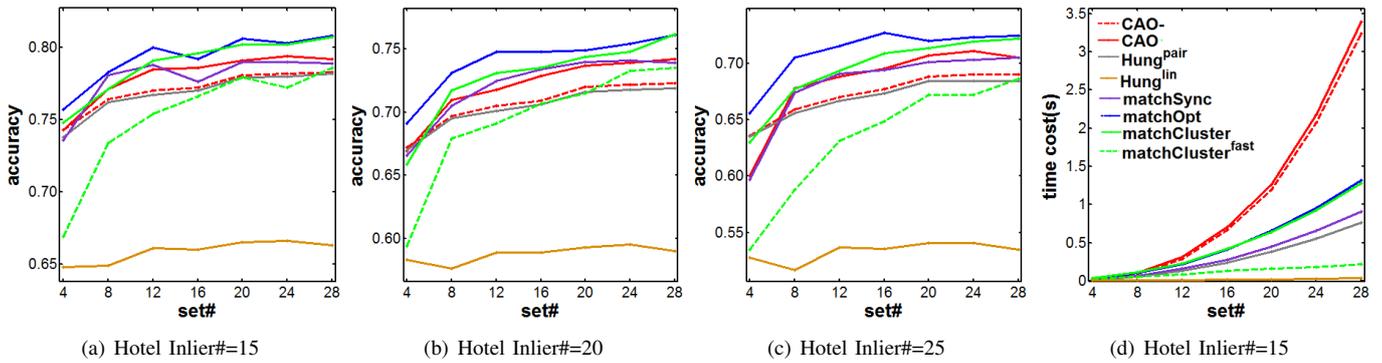

\centering
\subfigure[Hotel Inlier\#=15]{\label{fig:hotel_inlier=15_acc}
\includegraphics[width=0.245\textwidth]{{{pic/performance_cmu_hotel_RRWM_RRWM_respect_formal_inlier=15_acc}}}}\hspace{-5pt}
\subfigure[Hotel Inlier\#=20]{\label{fig:hotel_inlier=20_acc}
\includegraphics[width=0.245\textwidth]{{{pic/performance_cmu_hotel_RRWM_RRWM_respect_formal_inlier=20_acc}}}}\hspace{-5pt}
\subfigure[Hotel Inlier\#=25]{\label{fig:hotel_inlier=25_acc}
\includegraphics[width=0.245\textwidth]{{{pic/performance_cmu_hotel_RRWM_RRWM_respect_formal_inlier=25_acc}}}}\hspace{-5pt}
\subfigure[Hotel Inlier\#=15]{\label{fig:hotel_inlier=15_tim}
\includegraphics[width=0.245\textwidth]{{{pic/performance_cmu_hotel_RRWM_RRWM_respect_formal_inlier=15_tim}}}}\vspace{-5pt}
\caption{Matching accuracy and run-time under different inlier \# settings, as the number of feature sets vary on `Hotel' of CMU motion sequence.}
\label{fig:hotel}
\end{figure*}

\begin{figure}[t]
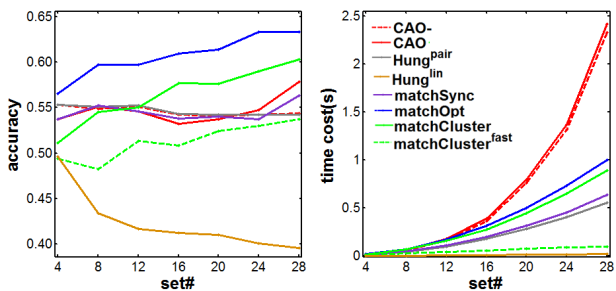

\centering
\subfigure{\label{fig:face_acc}
\includegraphics[width=0.22\textwidth]{{{pic/performance_object_face_RRWM_RRWM_respect_formal_all_acc}}}}\hspace{-5pt}
\subfigure{\label{fig:face_time}
\includegraphics[width=0.22\textwidth]{{{pic/performance_object_face_RRWM_RRWM_respect_formal_all_tim}}}}\vspace{-5pt}
\caption{Results on `Face' of Willow-ObjectClass as an average of 20 trials.}
\label{fig:face}
\end{figure}

\begin{figure*}[t]
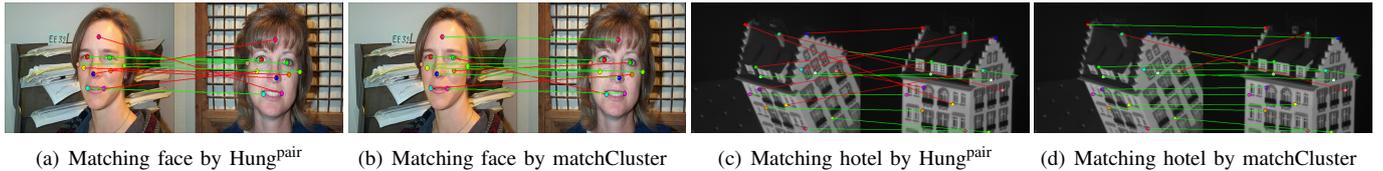

\centering
\subfigure[Matching face by Hung$^\text{pair}$]{\label{fig:random_grh_cnt_acc1}
\includegraphics[width=0.245\textwidth]{{{pic/object_face_image_0172_image_0176_raw_RRWM_RRWM_acc=0.5}}}}\hspace{-5pt}
\subfigure[Matching face by matchCluster]{\label{fig:random_grh_cnt_acc2}
\includegraphics[width=0.245\textwidth]{{{pic/object_face_image_0172_image_0176_ICPR_RRWM_RRWM_acc=0.7}}}}\hspace{-5pt}
\subfigure[Matching hotel by Hung$^\text{pair}$]{\label{fig:random_grh_cnt_acc3}
\includegraphics[width=0.245\textwidth]{{{pic/cmu_hotel_hotel1_hotel73_raw_RRWM_RRWM_acc=0.46667}}}}\hspace{-5pt}
\subfigure[Matching hotel by matchCluster]{\label{fig:random_grh_cnt_tim1}
\includegraphics[width=0.245\textwidth]{{{pic/cmu_hotel_hotel1_hotel73_ICPR_RRWM_RRWM_acc=0.86667}}}}\vspace{-5pt}
\caption{Illustration of the feature matching results on the face (Willow-ObjectClass) and hotel (CMU motion sequence) by the baseline pairwise Hungarian method for each pair of feature sets independently, and our method matchCluster. Red (green) lines indicate wrong (correct) correspondences.}
\label{fig:match_illustration}
\end{figure*}

The matching accuracy and run-time on the synthetic random tests are plotted in Fig.\ref{fig:random_vary_noise} and Fig.\ref{fig:vary_graph_cnt}. The former concerns the performance as a function of different disturbances, and the latter concerns the performance by changing the number of feature sets. The matching accuracy and run-time on the CMU Hotel sequence and Willow-ObjectClass Face is illustrated in Fig.\ref{fig:hotel} and Fig.\ref{fig:face}, respectively. Fig.\ref{fig:match_illustration} plots the visual matching illustration for the face and hotel image collections.

We make several observations based on the results:
\begin{enumerate}
  \item The synthetic tests suggest our methods (in green) are competitive especially in the presence of many outliers;
  \item In real image tests, matchOpt performs even better than ours, while ours perform competitively against other state-of-the-arts. No outlier is added in real image tests;
  \item matchCluster outperforms its variant matchCluster$^{\text{fast}}$ especially in real image tests. While in the synthetic test, the accuracy margin is much reduced;
  \item Though being most fast, while Hung$^{\text{lin}}$ always obtain the worst accuracy. This is reasonable as the information across feature sets are not fully explored by Hung$^{\text{lin}}$.
\end{enumerate}
Our analysis focuses on the first three bullets as 4) is obvious:

For 1) and 2), this perhaps suggests our method is more suited in two cases: i) when the deformation noises associated with the feature sets are evenly distributed around the template feature set -- that is the setting of our synthetic test; ii) when there are many outliers. For 3), we think the reason is that the real images are more heterogeneous which results in initial pairwise matchings with varying qualities. While in our synthetic tests, the noises are evenly distributed on the disturbed feature sets, thus their pairwise matching quality are relatively homogenous. This fact reduces the improvement space for adaptively deciding the updating order.

\section{Conclusion and future work}
We propose a novel method for matching feature sets with two advantages: i) the consistency of correspondences of the multiple feature sets is satisfied; ii) the global information of all feature sets is jointly explored. We show that given both simulated feature sets and real image sequences, the proposed method performs competitively with state-of-the-arts.

Future work involves designing and evaluating cost functions such as the K-nearest neighbor distance from the cluster. We will also apply our method to matching graphs by approximately extracting node signature e.g. the distribution of the length of the shortest paths from the node to other nodes.

\textbf{Acknowledgement} This work is supported by STCSM (15JC1401700, 14XD1402100, 13511504501).


\begin{thebibliography}{1}
\bibitem{ScottPRSL93}
G. Scott and H. Longuet-Higgins,
\newblock An algorithm for associating the features of two images,
\newblock in \emph{Proc. the Royal Society of London. Series B: Biological Sciences}, 1991.

\bibitem{ZengECCV12}
Z. Zeng, T.-H. Chan, K. Jia and D. Xu,
\newblock Finding correspondence from multiple images via sparse and low-rank decomposition,
\newblock in {\em ECCV}, 2012.

\bibitem{YuPR16}
J.-G. Yu, G.-S. Xia, A. Samal and J. Tian
\newblock Globally Consistent Correspondence of Multiple Feature Sets Using, Proximal Gauss-Seidel Relaxation,
\newblock in {\em Pattern Recognition}, 2016.

\bibitem{ConteIJPRAI04}
D.~Conte, P.~Foggia, C.~Sansone, and M.~Vento.
\newblock Thirty years of graph matching in pattern recognition.
\newblock {\em IJPRAI}, 2004.

\bibitem{ChoICCV13}
M.~Cho, K.~Alahari, and J.~Ponce.
\newblock Learning graphs to match.
\newblock In {\em ICCV}, 2013.

\bibitem{FoggiaIJPRAI14}
P.~Foggia, G.~Percannella, and M.~Vento.
\newblock Graph matching and learning in pattern recognition in the last 10 years.
\newblock {\em IJPRAI}, 33(1), 2014.

\bibitem{EppsteinSODA95}
D.~Eppstein,
\newblock Subgraph isomorphism in planar graphs and related problems,
\newblock in \emph{SODA}, 1995.

\bibitem{PachauriNIPS13}
D.~Pachauri, R.~Kondor, and S.~Vikas.
\newblock Solving the multi-way matching problem by permutation
  synchronization.
\newblock In {\em NIPS}, 2013.

\bibitem{ChenIVC92}
Y. Chen and G. Medioni,
\newblock Object modeling by registration of multiple range images,
\newblock {\em IVC}, 1992.

\bibitem{ZhangIJCV94}
Z. Zhang,
\newblock Iterative point matching for registration of free-form curves and surfaces,
\newblock {\em IJCV}, 1994.

\bibitem{LeordeanuIJCV12}
M.~Leordeanu, R.~Sukthankar, and M.~Hebert.
\newblock Unsupervised learning for graph matching.
\newblock {\em Int. J. Comput. Vis.}, pages 28--45, 2012.

\bibitem{ChuiCVIU03}
H.~Chui and A.~Rangarajan,
\newblock A new point matching algorithm for non-rigid registration,
\newblock {\em CVIU}, 2003.

\bibitem{BunkePAMI99}
H.~Bunke,
\newblock Error correcting graph matching: on the influence of the underlying cost function,
\newblock {\em PAMI}, 21(9), 1999.

\bibitem{YanAAAI15P}
J.~Yan, J.~Wang, H.~Zha, X.~Yang, and S.~Chu,
\newblock Multi-view point registration via alternating optimization,
\newblock In {\em AAAI}, 2015.

\bibitem{AflaloPNAS15}
Y.~Aflalo, A.~Bronstein, and R.~Kimmel,
\newblock On convex relaxation of graph isomorphism,
\newblock {\em PNAS}, 112(10):2942--2947, 2015.

\bibitem{WongPAMI85}
A.~Wong and M.~You,
\newblock Entropy and distance of random graphs with application to structural pattern recognition,
\newblock \emph{PAMI}, 1985.

\bibitem{LeordeanuICCV11}
M.~Leordeanu, A.~Zanfir, and C.~Sminchisescu.
\newblock Semi-supervised learning and optimization for hypergraph matching.
\newblock In {\em ICCV}, 2011.

\bibitem{ChoCVPR15}
M.~Cho, S.~Kwak, C.~Schmid, and J.~Ponce,
\newblock Unsupervised object discovery and localization in the wild: Part-based matching with bottom-up region proposals,
\newblock In {\em CVPR} 2015.

\bibitem{MacielPAMI03}
J. Maciel and J.P. Costeira,
\newblock A global solution to sparse correspondence problems,
\newblock PAMI, 2003.

\bibitem{ZhouICCV15}
X.~Zhou, M.~Zhu, and K.~Daniilidis,
\newblock Multi-image matching via fast alternating minimization,
\newblock In {\em ICCV}, 2015.

\bibitem{LoiolaEJOR07}
E.~M. Loiola and N.~Abreu and P.~O. Boaventura-Netto and P.~Hahn and T.~Querido,
\newblock A survey for the quadratic assignment problem.
\newblock {\em EJOR}, pages 657--90, 2007.

\bibitem{YanCVPR15}
J.~Yan, C.~Zhang, H.~Zha, W.~Liu, X.~Yang, and S.~Chu,
\newblock Discrete hyper-graph matching.
\newblock In {\em CVPR}, 2015.

\bibitem{YanICCV13}
J.~Yan, Y.~Tian, H.~Zha, X.~Yang, Y.~Zhang and S.~Chu,
\newblock Joint optimization for consistent multiple graph matching.
\newblock In {\em ICCV}, 2013.

\bibitem{YanTIP15}
J.~Yan, J.~Wang, H.~Zha, and X.~Yang,
\newblock Consistency-driven alternating optimization for multigraph matching: A unified approach.
\newblock {\em TIP}, 2015.

\bibitem{munkres1957algorithms}
J.~Munkres,
\newblock Algorithms for the assignment and transportation problems,
\newblock  \emph{Journal of the Society for Industrial \& Applied Mathematics}, vol.~5, no.~1, pp. 32--38, 1957.

\bibitem{JonkerComputing87}
R.~Jonker and A.~Volgenant
\newblock A shortest augmenting path algorithm for dense and sparse linear assignment problems,
\newblock \emph{Computing}, vol.~38, no.~4, pp. 325--340, 1987.

\bibitem{NgocCVPR15}
Q.~Ngoc, A.~Gautier, and M.~Hein,
\newblock A flexible tensor block coo rd inate ascent scheme for hypergraph matching,
\newblock In {\em CVPR}, 2015.


\bibitem{GoldPAMI96}
S.~Gold and A.~Rangarajan,
\newblock A graduated assignment algorithm for graph matching,
\newblock \emph{IEEE Transaction on PAMI}, 1996.

\bibitem{LeePR93}
C.-H. Lee and A. Joshi,
\newblock Correspondence problem in image sequence analysis
\newblock \emph{Pattern Recognition},  26(1) (1993) 47--61.

\bibitem{TianECCV12}
Y.~Tian, J.~Yan, H.~Zhang, Y.~Zhang, X.~Yang, and H.~Zha,
\newblock On the convergence of graph matching: Graduated assignment revisited,
\newblock In {\em ECCV}, 2012.

\bibitem{YanPAMI16}
J.~Yan, M.~Cho, H.~Zha, X.~Yang, and S.~Chu,
\newblock Multi-graph matching via affinity optimization with graduated,
  consistency regularization.
\newblock {\em TPAMI}, 2016.

\bibitem{LeordeanuNIPS09}
M.~Leordeanu, M.~Hebert, and R.~Sukthankar.
\newblock An integer projected fixed point method for graph matching and map inference,
\newblock In {\em NIPS}, 2009.

\bibitem{YanECCV14}
J.~Yan, Y.~Li, W.~Liu, H.~Zha, X.~Yang, and S.~Chu,
\newblock Graduated consistency-regularized optimization for multi-graph matching,
\newblock In {\em ECCV}, 2014.

\bibitem{LoweICCV99}
D. Lowe,
\newblock Object recognition from local scale-invariant features,
\newblock In {\em ICCV}, 1999.

\bibitem{YanACCV09}
J. Yan, Y. Li, E. Zheng and Y. Liu,
\newblock An accelerated human motion tracking system based on voxel reconstruction under complex environments,
\newblock In {\em ACCV}, 2009.

\bibitem{ShiCVPR16}
X. Shi and H. Ling and W. Hu and J. Xing and Y. Zhang,
\newblock Tensor Power Iteration for Multi-Graph Matching,
\newblock In {\em CVPR}, 2016.

\bibitem{ShiCVPR13}
X. Shi and H. Ling and J. Xing and W. Hu,
\newblock Multi-target Tracking by Rank-1 Tensor Approximation,
\newblock In {\em CVPR}, 2013.


\bibitem{MyronenkoPAMI10}
A. Myronenko and X. Song,
\newblock Point set registration: Coherent point drift,
\newblock {\em TPAMI}, 2010.

\bibitem{luks1982isomorphism}
E.~M. Luks
\newblock Isomorphism of graphs of bounded valence can be tested in polynomial time,
\newblock \emph{Journal of Computer and System Sciences}, 1982.

\bibitem{aho1989code}
A.~V. Aho, M.~Ganapathi, and S.~W. Tjiang
\newblock Code generation using tree matching and dynamic programming,
\newblock \emph{TOPLAS}, 1989.

\bibitem{neuhaus2005self}
M.~Neuhaus and H.~Bunke.
\newblock Self-organizing maps for learning the edit costs in graph matching,
\newblock {\em IEEE Transactions on Systems, Man, and Cybernetics, Part B:
  Cybernetics}, 35(3):503--514, 2005.

\bibitem{neuhaus2007automatic}
M.~Neuhaus and H.~Bunke
\newblock Automatic learning of cost functions for graph edit distance,
\newblock {\em Information Sciences}, 177(1):239--247, 2007.

\bibitem{niu2015applying}
Z. Niu and R. Martinand M. Sabin and F. Langbein and H. Bucklow
\newblock Applying database optimization technologies to feature recognition in CAD,
\newblock {\em Computer-aided design and applications}, 12(3):373--382, 2015.

\bibitem{NiuCAD15}
Z. Niu and R. Martinand M. Sabin and F. Langbein and M. Sabin
\newblock Rapidly finding CAD features using database optimization,
\newblock {\em Computer-aided design and applications}, 69:35--50, 2015.


\bibitem{YanYanPAMI16}
Y. Yan and E. Ricci and R. Subramanian and G. Liu and O. Lanz and N. Sebe
\newblock LA Multi-task Learning Framework for Head Pose Estimation under Target Motion,
\newblock {\em PAMI}, 2016.
\end{thebibliography}
\end{document}